\documentclass[conference]{IEEEtran}
\usepackage{amssymb,amsfonts}
\usepackage{times}
\usepackage{enumitem}   
\usepackage{bbm}
\usepackage{lipsum}
\usepackage{comment}
\usepackage{array}
\usepackage{mathtools} 
\usepackage{algorithm}
\usepackage{amsmath,amsthm,bm}
\usepackage{soul,xcolor}
\usepackage{mdwmath}
\usepackage{mdwtab}
\usepackage{mathrsfs}
\usepackage{subcaption}
\usepackage{amsmath,amssymb,amsfonts}
\usepackage{algorithmic}
\usepackage{graphicx}
\usepackage{textcomp}

\usepackage[utf8]{inputenc}
\usepackage[english]{babel}
\newtheorem{theorem}{Theorem}

\newtheorem{lemma}{Lemma}

\def\BibTeX{{\rm B\kern-.05em{\sc i\kern-.025em b}\kern-.08em
    T\kern-.1667em\lower.7ex\hbox{E}\kern-.125emX}}
\begin{document}

\title{Certified Robustness of Graph Classification against Topology Attack with Randomized Smoothing
}
\author{\IEEEauthorblockN{Zhidong Gao, Rui Hu, Yanmin Gong}
\IEEEauthorblockA{Department of Electrical and Computer Engineering, University of Texas at San Antonio, San Antonio, TX 78249}
}
\maketitle

\begin{abstract}
Graph classification has practical applications in diverse fields. Recent studies show that graph-based machine learning models are especially vulnerable to adversarial perturbations due to the non i.i.d nature of graph data. By adding or deleting a small number of edges in the graph, adversaries could greatly change the graph label predicted by a graph classification model. In this work, we propose to
build a smoothed graph classification model with certified robustness guarantee. We have proven that the resulting graph classification model would output the same prediction for a graph under $l_0$ bounded adversarial perturbation. We also evaluate the effectiveness of our approach under graph convolutional network (GCN) based multi-class graph classification model.
\end{abstract}

\section{Introduction}
Graph has strong capability to represent natural relational data such as citation network, bio-medical molecular, social network etc., which provides wide opportunities to utilize graph based machine learning models. 

Graph classification, or the problem of assigning labels to a graph in a dataset, has found practical applications in diverse fields, including malware detection\cite{chau2011polonium}, learning molecular fingerprints \cite{duvenaud2015convolutional}, and anticancer hyperfood prediction\cite{gonzalez2020graph}. For example, in chemoinformatics, graphs can be used to represent molecules, with nodes signifying atoms and edges denoting chemical bonds of atom pairs, and a label of a graph given by a graph classification algorithm could denote molecular properties such as the anti-cancer activity, solubility, or toxicity. While there exist many classic approaches, graph neural networks (GNNs) represent the state-of-the-art in frequently applied tasks on graph data and have attracted great attention recently \cite{kipf2017semi,hamilton2017inductive,velickovic2018graph}.

Despite of their remarkable performance, GNNs have been shown to be vulnerable to adversarial attacks \cite{zugner2018adversarial,zugner_adversarial_2019}, i.e., deliberately designed and small perturbations in the graph structure or node features of a graph could lead to drastically degraded performance of a GNN model. Such observation poses great challenges in applying GNNs to real-world applications, especially safety-critical scenarios such as healthcare and transportation. Therefore, ensuring the robustness of GNN models is of significant importance to the wide applicability of these models.

Compared with traditional machine learning models, it is especially challenging to ensure the robustness of GNN models due to the non-i.i.d. nature of graph data. Specifically, the adversarial effects of an attack against a node or edge can propagate to other nodes or edges via the graph structure, making the space of possible attacks very large. Until now, there are only very few mechanisms that can effectively defend against the adversarial attacks in GNNs \cite{zhu2019robust}. Most proposed mechanisms are \emph{best-effort} and heuristic methods for training GNN models intended to be robust to adversarial attacks, and they are most likely to fail given suitably powerful adversaries as observed from the robustness research in traditional machine learning models. 

On the other hand, certifiable robustness ensures that a classifier whose prediction at any point is verifiably constant within some set around the point. By offering a theoretically justified guarantee of robustness, adversarial attacks can be provably defended. The research line of providing certifiable robustness for GNN models has emerged in very recent studies \cite{bojchevski2019certifiable,zugner2019certifiable}. However, these studies tend to be tied to internal GNN model details, such as the types of aggregation and activation functions, and cannot easily generalize across different types of GNN models. Moreover, their computational complexities are often very high (e.g., solving an NP-hard problem) and do not immediately scale to large GNN models.

Providing robustness guarantee for graph classification model in these applications is an important but unsolved problem. In this paper, we focus on providing robustness guarantee for graph classification models based on randomized smoothing\cite{cohen2019certified,lecuyer2019certified,li2019certified}, which turns any base classifier into a robust classifier via adding random noise (usually Gaussian) to the input data. There are several challenges originate from the inherent nature of graph data though, including how to add commonly used Gaussian noises to the discrete graph structure and how to provide discrete $l_0$ certification instead of $l_2$ in graph data. We address the above challenges by using discrete random noise sampled from Bernoulli distribution and deriving the analytical bound for the robustness guarantee of the smoothed GNN classifier in the $l_0$ norm metric. 

The contributions of this paper are summarized as follows:
    
    
    

\begin{itemize}
    \item We propose a certifiable robustness method for graph classification models that can achieve both scalability and general applicability.

    \item We propose to use Bernoulli distribution as the random noise to perturb the graph data and smoothen the decision boundaries of the classic GNN models by ensemble prediction.
    
    \item We evaluate our approach based on  graph  convolutional  network (GCN), and empirically demonstrate the effectiveness of our approach for multi-class graph classification.
\end{itemize}

The rest of the paper is organized as follows. First, the related works is introduced in Section~\ref{sec:related}. Then, our proposed method to provide certifiable robustness of graph classification models based on randomized smoothing is illustrated in Section~\ref{sec:our_approach}. Next, experimental results are given in Section~\ref{sec:experiment}. Finally, Section~\ref{sec:conclusion} makes the conclusion.


\section{Related Work}\label{sec:related}
Our work is related to two categories of recent research: graph classification and graph adversarial attacks.

\subsection{Graph classification}

There are several approaches of graph classification, including kernel-based approaches which measure similarity between graphs\cite{ying2018hierarchical,kriege2016valid}, and neural network approaches which leverage machine learning framework to learn graph features\cite{scarselli2008graph}. Inspired by recent progress of Convolutional Neural Network (CNN), two approaches, convolutional approach\cite{kipf2017semi,velivckovic2017graph,hamilton2017inductive} and pooling approach\cite{ying2018hierarchical,gao2019graph,lee2019self,ma2019graph} are proposed with the aim to generalize the convolution and pooling operation in graph neural network. Among these approaches, graph neural network approach, especially GCN, has been widely used since it provides state-of-the-art performance in many cases.


\subsection{Attack and Defense in Graph-based Learning}
Machine learning models are vulnerable to attacks. Two of the main attack scenarios are evasion attacks \cite{goodfellow2014explaining,biggio2013evasion} and poisoning attacks \cite{mei2015using,biggio2013security,nelson2008exploiting,steinhardt2017certified,munoz2017towards}, focusing on test time and training time, respectively. Compared to non-graph data, attacks and defenses on graph data are less investigated. Previous attacks on graph-based learning are mainly focusing on transductive setting, and thus are mostly poisoning attacks \cite{liu2019unified,ijcai2019-674}. Little evasion attacks on graph-based learning are investigated \cite{ijcai2019-669}. With the development of new inductive learning algorithms on graph data, the risk of evasion attacks increases.  Evasion attacks are often instantiated by adversarial examples, which are crafted by making small, often imperceptible perturbations to legitimate inputs, with the goal of misleading a learned classifier to misclassify the resulting adversarial inputs. Note that adversarial examples generated from one model usually are also harmful for other model, known as transferability \cite{tramer2017space}. In this paper, we investigate how we could improve the robustness of graph-classification algorithms against evasion attacks. 

\section{Certified Robustness for graph classification}\label{sec:our_approach}
In this section, we will introduce notations and our approach to provide certified robustness guarantee for graph classification model.
\subsection{Notations and Graph Classification Model}\label{sec:Notations}

Given a set of graphs $\mathcal{D}=\{(\mathcal{G}_{1},\mathnormal{c}_{1}), (\mathcal{G}_{2},\mathnormal{c}_{2}),\cdots, \}$, where the number of nodes may differ in each graph. For an attributed graph $\mathcal{G} = (A,X)$, we have adjacency matrix $A\in \{0,1\}^{n\times n}$ and feature matrix $X \in \{0,1\}^{n \times d}$, where $n$ denotes the number of nodes in corresponding graph and $d$ is the number of node features. For each graph, it has related label $c \in \mathcal{C}$, where $\mathcal{C}$ denotes the set of labels. The goal of graph classification is to learn a function $f: \mathbb{G} \rightarrow \mathcal{C} $, where $\mathbb{G}$ is input space of graphs.
\subsection{Attack model}
We consider the attack scenario where an attacker can only change the topology of graph $\mathcal{G}$ by removing existing edges or adding new edges into the original edge set, forming a new graph $\mathcal{\widetilde{G}}$. For simplicity, we use a binary vector \textbf{x}\textbf{$_\mathcal{G} \in \{0,1\}^N$} to represent the flattened adjacency matrix A, where $N$ denotes the length of vector. Then, the adversarial perturbation vector $\bm{\epsilon}$ is introduced to model whether a corresponding edge is perturbed or not, and $\bm{\epsilon}_i$ denotes the $i$-th element of $\bm{\epsilon}$. Specifically, $\bm{\epsilon}_i=1$ if the corresponding edge is perturbed, otherwise, $\bm{\epsilon}_i=0$. Formally, given the binary vector \textbf{x}\textbf{$_\mathcal{G}$} of original graph $\mathcal{G}$, by perturbing the edges according to the edge perturbation vector $\bm{\epsilon}$, the new binary vector  \textbf{$\widetilde{\textbf{x}}$}\textbf{$_\mathcal{G}$} of perturbed graph can be represent as: 
\begin{equation}\label{eq:attack1}
\widetilde{\textbf{x}}\textbf{$_\mathcal{G}$} = \textbf{x}\textbf{$_\mathcal{G}$} \oplus \bm{\epsilon}, 
\end{equation}
where the operator $\oplus$ denotes the ``exclusive or'' binary operation.

\subsection{Achieving Certifiable Robustness with Randomized Smoothing}

With randomized smoothing, we construct a new, ``smoothed'' classifier $g$ from an arbitrary base GNN classifier $f$. When queried at a specific unlabeled graph $\mathcal{G}$, the smoothed classifier $g$ returns whichever label the base classifier $f$ is mostly likely to return when the topology of graph $\mathcal{G}$ is perturbed by certain random noises:
\begin{equation}\label{eq:eps}
  \Pr(\epsilon_i) =
    \begin{cases}
      \beta & \epsilon_i = 0\\
      1 - \beta & \epsilon_i = 1
    \end{cases},
    \forall i \in \{1,2,\cdots,N \}
\end{equation}
where $i$ is $i$-th element of binary vector $\bm{\epsilon}$. In our scenario each element in the binary vector has probability $\beta$ to keep unchanged and has probability $1-\beta$ to flip the structure. Then, the smoothed classifier is defined as:
\begin{equation}\label{eq:g}
g(\textbf{$\textbf{x}_\mathcal{G}$}) = \underset{c \in \mathcal{C}}{\operatorname{argmax}}\;\Pr (f(\textbf{$\textbf{x}_\mathcal{G}$} \oplus \bm{\epsilon}) = c)
\end{equation}
where $c$ is the class which has largest probability measure under noise perturbation $\bm{\epsilon}$. 
Certifying robustness against any adversarial attack is to certify that $g(\textbf{$\textbf{x}_\mathcal{G}$} \oplus \bm{\delta}) = c$ 
for all $\bm{\|{\delta}\|_{0}} = L$, 
where $L$ denotes certified perturbation size.

Next we will introduce how to derive the certified perturbation size of smoothed classifier $g$. Given the smoothed classifier $g$ defined in (\ref{eq:g}), one can certify the model's output against adversarial perturbation within range $L$. Formally, we have the following theorem:

\begin{theorem}\label{theorem_1}
Given a graph represented with \textbf{x}\textbf{$_\mathcal{G} \in \{0,1\}^N$}, a graph classification model $f: \mathbb{G} \rightarrow \mathcal{C} $, and a smoothed classifier $g$ defined in (\ref{eq:g}). 
Suppose $c_A \in \mathcal{C}$ and there exists $\underline{p_A}, \overline{p_B} \in [0,1]$ such that
\begin{equation}\label{eq:cond}
\Pr (f(\textbf{$\textbf{x}_\mathcal{G}$} \oplus \bm{\epsilon}) = c_A) \geq \underline{p_A} \geq  \overline{p_B} \geq \underset{c \neq c_A}{\operatorname{max }}\;\Pr (f(\textbf{$\textbf{x}_\mathcal{G}$} \oplus \bm{\epsilon}) = c).
\end{equation}
Here, $\underline{p_A}$ denotes the lower bound probability of the most probable class $c_A$  and $\overline{p_B}$  denotes the upper bound probability of the ``runner-up'' class of $f$ under random noise $\epsilon$. 
Then, $g(\textbf{$\textbf{x}_\mathcal{G}$} \oplus \bm{\epsilon}) = c_A$ for all  $\|{\delta}\|_0 < L$, where $L$ is certified perturbation size and can be calculated by solving the follow optimization problem:
\begin{equation}\label{eq:opti}
\begin{aligned}
L = & \operatorname{argmax}\;l\\
s.t. & \|{\bm{\delta}}\|_0 =l,\\              
&\sum_{k=\mu{2}}^{\mu_{1}-1}\Pr (\textbf{$\textbf{x}_{\mathcal{G} }$} \oplus \bm{\epsilon}  \oplus \bm{\delta} \in \mathcal{H}_{k}) + \\
&(\underline{p_{A}}-\sum_{k=1}^{\mu_{1}-1}\Pr(\textbf{x}_\mathcal{G}
\oplus \bm{\epsilon}  \in \mathcal{H}_{k}))\cdot\frac{\Pr(\textbf{$\textbf{x}_{\mathcal{G} }$} 
\oplus \bm{\epsilon}  \oplus \bm{\delta} \in \mathcal{H}_{\mu_{1}})}{\Pr\left(\textbf{x}_{\mathcal{G}}
\oplus \bm{\epsilon}  \in \mathcal{H}_{\mu_1}\right)} > \\
&(\overline{p_{B}}-\sum_{k=1}^{\mu_{2}-1}\Pr(\textbf{x}_\mathcal{G} \oplus \bm{\epsilon}  \in 
\mathcal{H}_{k}))
\cdot\frac{\Pr(\textbf{$\textbf{x}_{\mathcal{G} }$} \oplus \bm{\epsilon}  \oplus \bm{\delta} \in 
\mathcal{H}_{\mu_{2}})}{\Pr\left(\textbf{x}_{\mathcal{G}}  \oplus \bm{\epsilon}  \in \mathcal{H}_{\mu_2}\right)}
\end{aligned}
\end{equation}
The region $\mathcal{H}(e)$ and density ratio $h(e)$ are defined as \cite{jia2020certified}:
\begin{equation}
    \mathcal{H}(e) = \{ \textbf{z} \in \{0,1\}^{N } \}: \frac{\Pr(\textbf{x}_{\mathcal{G} } \oplus \bm{\epsilon}  =z)}{\Pr(\textbf{x}_{\mathcal{G}}  \oplus \bm{\epsilon}  \oplus \bm{\delta} = z)}
\end{equation}
\begin{equation}
    h(e) = (\frac{\beta}{1-\beta})^e
\end{equation}
where $e = -N , -N +1, \cdots, N -1, N $. The region $\mathcal{H}(-N ), \mathcal{H}(-N +1),\cdots, \mathcal{H}(N )$ is ranked in an ascending order according to the density ratio $h(-N ),h(-N +1),\cdots,h(N )$. And we denote them as $\mathcal{H}_1,\mathcal{H}_2,\cdots,\mathcal{H}_{2N +1}$ according to their orders. We define $\mu_1, \mu_2$ as follows:
\begin{equation}
\begin{split}
    &\mu_1 = \underset{\mu' \in \{1,2,\cdots,2N +1\}}{\operatorname{argmin}} \mu', s.t. \sum_{k=1}^{\mu'}\Pr(\textbf{x}_{\mathcal{G} } \oplus \bm{\epsilon}  \in \mathcal{H}_k) \geq \underline{p_A}\\
    &\mu_2 = \underset{\mu' \in \{1,2,\cdots,2N +1\}}{\operatorname{argmin}} \mu', s.t. \sum_{k=1}^{\mu'}\Pr(\textbf{x}_{\mathcal{G} } \oplus \bm{\epsilon}  \in \mathcal{H}_k) \geq \overline{p_B}
\end{split}
\end{equation}
\end{theorem}

\begin{IEEEproof}
Here, we first restate the Neyman-Pearson Lemma under discrete space \cite{jia2020certified}, and then provide the proof for Theorem~\ref{theorem_1}.

\begin{lemma}\label{lemma_1}
Assume $X$ and $Y$ are two random variables in the discrete space $\{0,1\}^n$ with probability distribution $\Pr(X)$ and $\Pr(Y)$, respectively. Let $\psi: \{0,1\}^n \rightarrow \{0,1\}$ be a random  or deterministic function. Let $T_1 = \{\textbf{z} \in \{0,1\}^N: \frac{\Pr(X=\text{z})}{\Pr(Y=\text{z})}\}$ and $T_2 = \{\textbf{z} \in \{0,1\}^n: \frac{\Pr(X=\text{z})}{\Pr(Y=\text{z})}\}$ for some $t > 0$. Assume  $T_3 \subseteq T_2$ and $T = T_1 \cup T_3$. If $\Pr(\psi(X) = 1) \geq \Pr(X \in T)$, then $\Pr(\psi(Y) = 1) \geq \Pr(Y \in T)$. If $\Pr(\psi(X) = 1) \leq \Pr(X \in T)$, then $\Pr(\psi(Y) = 1) \leq \Pr(Y \in T)$.
\end{lemma}

We first define two random variables as follow:
\begin{equation}
\begin{split}
&X = \textbf{x}_{\mathcal{G} } \oplus \bm{\epsilon } \\
&Y = \textbf{x}_{\mathcal{G} } \oplus \bm{\delta} \oplus \bm{\epsilon }
\end{split}
\end{equation}
which represent the random samples after adding noise to the binary vector $\textbf{x}_{\mathcal{G} }$ 
and $\textbf{x}_{\mathcal{G} \oplus \bm{\delta}}$. Our goal is to find the maximum perturbation size $\|\bm{\delta}\|_0$ such that following condition hold:
\begin{equation}
    \Pr(f(Y)=c_A) > \Pr(f(Y)=c_B)
\end{equation}
We first define two regions $\mathcal{Q}_1,\mathcal{Q}_2$ such that $\Pr(X\in \mathcal{Q}_1) = \underline{p_A}, \Pr(X \in \mathcal{Q}_2) = \overline{p_B}$. 
Specifically, we gradually add the region $\mathcal{H}_{1},\mathcal{H}_{2},\cdots,\mathcal{H}_{2N+1}$ 
to the $\mathcal{Q}_{1},\mathcal{Q}_{2}$ 
up to $\Pr(X \in \mathcal{Q}_1) = \underline{p_A}, \;\Pr(X \in \mathcal{Q}_2) = \overline{p_B}$. 
In particular, we define $\mu_1,\mu_2$ as:
\begin{equation}
\begin{split}
    &\mu_1 = \underset{\mu' \in \{1,2,\cdots,2N +1\}}{\operatorname{argmin}} \mu', s.t. \sum_{k=1}^{\mu'}\Pr(\textbf{x}_{\mathcal{G} } \oplus \epsilon  \in \mathcal{H}_k) \geq \underline{p_A}\\
    &\mu_2 = \underset{\mu' \in \{1,2,\cdots,2N +1\}}{\operatorname{argmin}} \mu', s.t. \sum_{k=1}^{\mu'}\Pr(\textbf{x}_{\mathcal{G} } \oplus \epsilon  \in \mathcal{H}_k) \geq \overline{p_B}
\end{split}
\end{equation}
Moreover, we define $\underline{\mathcal{H}_{\mu_{1}}},\underline{\mathcal{H}_{\mu_{2}}}$ as  any subregion of $\mathcal{H}_{\mu_{1}},\mathcal{H}_{\mu_{2}}$ such that:
\begin{equation}
\begin{split}
    &\Pr(X \in \underline{\mathcal{H}_{\mu_{1}}}) = \underline{p_A} - \sum_{k=1}^{\mu_1-1}\Pr(X \in \mathcal{H}_k)\\
    &\Pr(X \in \underline{\mathcal{H}_{\mu_{2}}}) = \overline{p_B} - \sum_{k=1}^{\mu_2-1}\Pr(X \in \mathcal{H}_k)
\end{split}
\end{equation}
Then, the region $\mathcal{Q}_1,\mathcal{Q}_2$ can be represent as:
\begin{equation}
\begin{split}
    &\mathcal{Q}_1 = \bigcup_{k=1}^{\mu_1-1}\mathcal{H}_k\cup\underline{\mathcal{H}_{\mu_1}}\\
    &\mathcal{Q}_2 = \bigcup_{k=1}^{\mu_2-1}\mathcal{H}_k\cup\underline{\mathcal{H}_{\mu_2}}
\end{split}
\end{equation}
Based on the condition of equation \ref{eq:cond}, we have:
\begin{equation}
\begin{split}
    &\Pr(f(X)=c_A) \geq \underline{p_A} = \Pr(X \in \mathcal{Q}_1)\\
    &\Pr(f(X)=c_B) \leq \overline{p_B} = \Pr(X \in \mathcal{Q}_2)
\end{split}
\end{equation}
Define the function $\psi(\textbf{z})=\mathbb{I}(f(\textbf{z})=c)$. Then, we have:
\begin{equation}
\begin{split}
    &\Pr(\psi(X)=c_A) = \Pr(f(X)=c_A) = \Pr(X \in \mathcal{Q}_1)\\
    &\Pr(\psi(X)=c_B) = \Pr(f(X)=c_B) = \Pr(X \in \mathcal{Q}_2)
\end{split}
\end{equation}
Moreover, we have $\frac{\Pr(X=\textbf{z})}{\Pr(Y=\textbf{z})}> h_{\mu_{1}}$, $\frac{\Pr(X=\textbf{z})}{\Pr(Y=\textbf{z})}> h_{\mu_{2}}$ if and only if $\textbf{z}\in\bigcup_{j=1}^{\mu_1-1}\mathcal{H}_j$, $\textbf{z}\in\bigcup_{j=1}^{\mu_2-1}\mathcal{H}_j$ 
separately. We also have $\frac{\Pr(X=\textbf{z})}{\Pr(Y=\textbf{z})} = h_{\mu_{1}}$, $\frac{\Pr(X=\textbf{z})}{\Pr(Y=\textbf{z})} =  h_{\mu_{2}}$ 
for any $\textbf{z}\in\underline{\mathcal{H}_{\mu_1}}$, $\textbf{z}\in\underline{\mathcal{H}_{\mu_2}}$. 
Note that $\mathcal{Q}_1 = \bigcup_{k=1}^{\mu_1-1}\mathcal{H}_k\cup\underline{\mathcal{H}_{\mu_1}}$ and $\mathcal{Q}_2 = \bigcup_{k=1}^{\mu_2-1}\mathcal{H}_k\cup\underline{\mathcal{H}_{\mu_2}}$, According to Neyman-Pearson Lemma mentioned before, we can obtain that:
\begin{equation}
    \begin{split}
        &\Pr(f(Y)=c_A) \geq \Pr(Y \in \mathcal{Q}_1)\\
        &\Pr(f(Y)=c_B) \leq \Pr(Y \in \mathcal{Q}_2)
    \end{split}
\end{equation}
To reach our goal, it's sufficient to have:
\begin{equation}
\begin{split}
    &\Pr(Y \in \mathcal{Q}_1) \geq \Pr(Y \in \mathcal{Q}_2) \Leftrightarrow\\
    &\Pr(Y \in \bigcup_{k=1}^{\mu_1-1}\mathcal{H}_k\cup\underline{\mathcal{H}_{\mu_1}})>\Pr(Y \in \bigcup_{k=1}^{\mu_2-1}\mathcal{H}_k\cup\underline{\mathcal{H}_{\mu_2}} \Leftrightarrow\\
    &\sum_{k=1}^{\mu_1-1}\Pr(Y \in \mathcal{H}_k) + (\overline{p_B} - \sum_{k=1}^{\mu_1-1}\Pr(X \in \mathcal{H}_k))/h_{\mu_{1}} > \\
    &\sum_{k=1}^{\mu_2-1}\Pr(Y \in \mathcal{H}_k) + (\underline{p_A} - \sum_{k=1}^{\mu_2-1}\Pr(X \in \mathcal{H}_k))/h_{\mu_{2}}
\end{split}
\end{equation}
\end{IEEEproof}
Next, we present a practical algorithm evaluating the $g(\textbf{x}_{\mathcal{G} })$ and provide the certified robustness guarantee of $g$ around $\textbf{x}_{\mathcal{G}}$. Given a base graph classifier $f$ and a graph $\mathcal{G} $, the evaluation includes two stages. Firstly, we need to identify the most probable class $c_A$. Then, we need to estimate the lower bound probability $\underline{p_A}$ and upper bound probability $\overline{p_B}$ under random noise corrupted input $\textbf{x}_{\mathcal{G} } \oplus \bm{\epsilon} $. 

Our approach is implemented based on Monte Carlo method. We first sample $M$ random noises and use correspondingly perturbed graphs as input. Denote the most frequently appeared class when querying the output as $c_A$. Formally, we have:
\begin{equation}
\begin{aligned}
    &c_A = & & \underset{c \in \mathcal{C}}{\operatorname{argmax}}\;\eta_c \\
    &s.t.  & & \eta_c = \sum_{m=1}^{M}\mathbb{I}(f(\textbf{x}_{\mathcal{G}_m } \oplus \bm{\epsilon} =c)),
\end{aligned}
\end{equation}
where $\eta_c$ is counter of class $c$ appeared during the sampling, and $\mathbb{I}$ is the indicator function.

Then, the lower-bound probability $\underline{p_A}$ can be estimated using one-sided Clopper-Pearson method:
\begin{equation}
    \underline{p_A} = LCB(\eta_c,M,1-\alpha),
\end{equation}
where $LCB(\cdot)$ denotes lower-confidence-bound function which returns one-sided lower confidence interval for the Binomial parameter $p$ such that $\eta_c\sim \text{Binomial}(M,p)$ with probability $1-\alpha$.

For the upper bound probability $\overline{p_B}$, it's very hard to give accurate maximum probability among remain classes if $|\mathcal{C}|>2$. In actual algorithm, we just take: $\overline{p_B} = 1- \underline{p_A}$. And in this situation, $\overline{p_B}<1-\underline{p_A}$ leads $1-\underline{p_A}<\underline{p_A}$, which is $\underline{p_A}>0.5$ in return. If $\underline{p_A}<=0.5$, the algorithm will abstain. 

To solve the optimization problem in equation (\ref{eq:opti}), the key is calculating two probabilities $\Pr(\textbf{$\textbf{x}_{\mathcal{G} }$} \oplus \bm{\epsilon}  \oplus \bm{\delta} \in \mathcal{H}(e))$ and $\Pr\left(\textbf{x}_\mathcal{G}  \oplus \bm{\epsilon}  \in \mathcal{H}(e)\right)$. Formally, we have:
\begin{equation}
    \Pr\left(\textbf{x}_\mathcal{G}  \oplus \bm{\epsilon}  \in \mathcal{H}(e)\right) = \sum_{k=max\{0,e\}}^{min\{N ,N +e\}} \beta^{N -(k-e)}(1-\beta)^{k-e}\cdot\theta(k),
\end{equation}
\begin{equation}
    \Pr\left(\textbf{x}_\mathcal{G}  \oplus \bm{\delta}  \oplus \bm{\epsilon}  \in \mathcal{H}(e)\right) = \sum_{k=max\{0,e\}}^{min\{N ,N +e\}} \beta^{N -k}(1-\beta)^{k}\cdot\theta(e,k),
\end{equation}
where $\theta(k)$ is:
\begin{equation}
    \theta(e,k) = 
    \begin{cases}
        0, &if\;(e+l)\;mod\;2\neq0\\
        0, &if\;2k-e<l\\
        \binom{N -k}{\frac{2k-e-l}{2}}\binom{l}{\frac{l-e}{2}}, &otherwise
    \end{cases}
\end{equation}
After we calculate the two probabilities, the optimization problem can be solved iteratively.
Our whole algorithm is summarized in Algorithm \ref{alg:certification}.

\begin{algorithm}[ht]
\caption{Certification and Prediction}
\label{alg:certification}
\begin{algorithmic}[1]
\REQUIRE $f$, $\textbf{x}_\mathcal{G} $, $\beta$, M, $\alpha$
\STATE counts $\longleftarrow$ SAMPLE($f$, $\beta$, $\textbf{x}_\mathcal{G} $, M);
\STATE $\hat{c}_A$, $\hat{c}_B$ $\longleftarrow$ top two classes in counts;
\STATE$\eta_A$, $\eta_B$ $\longleftarrow$ count [$\hat{c}_A$], count [$\hat{c}_B$];
\STATE$\underline{p_A}$ $\longleftarrow$ LCB($\eta_A$, M, $1-\alpha$);
\IF{BPV($\eta_A$, $\eta_A+\eta_B$, 0.5) $\leq$ $\alpha$ and $\underline{p_A}$ $>$ $\frac{1}{2}$}
    \STATE L = RADIUS($\underline{p_A}$, $\overline{p_B}$, $\textbf{x}_\mathcal{G} $);
    \RETURN ($\hat{c}_A$, L);
\ELSE
   \STATE ABSTAIN;
\ENDIF
\RETURN ABSTAIN or ($\hat{c}_A$, L)
\end{algorithmic}
\end{algorithm}

In our algorithm, the function BPV($\eta_A$, $\eta_A+\eta_B, p) \leq$ $\alpha$ returns the p-value of two-sided hypothesis test that $\eta_A \sim$ Binomial$(\eta_A+\eta_B,p)$. The function RADIUS solves equation (\ref{eq:opti}) and returns the certified perturbation size $L$. The function SAMPLE($f$, $\beta$, $\textbf{x}_\mathcal{G} $, M) randomly generates $M$ samples with noise control level $\beta$, gives prediction for each $\textbf{x}_\mathcal{G} \oplus \bm{\epsilon}$, 
and returns the frequency of each class.

\section{Performance Evaluation}\label{sec:experiment}
In this section, we evaluate the performance of our proposed certifiable robustness method. We first describe our experimental setup. We then demonstrate the effectiveness of our method and study the performance of our method under different scenarios.

\subsection{Experimental Setup}
We evaluate our method for graph classification on a synthetic dataset. The synthetic dataset consists of 480 graphs with different topologies, including cycle-type, star-type, wheel-type, lollipop-type, hypercube-type, grid-type, complete-type, and cicular ladder-type topology. Our goal is to classify these graphs into different categories of topology. We use the well-established GCN algorithm \cite{kipf2017semi} to train the base graph classifier. Specifically, we establish a two-layer GCN model and 
train it beforehand using 320 samples. After obtaining the trained GCN model, we implement our algorithm on it and retrain the model, and finally make predictions on 160 test samples. For the $m$-th testing sample, we can obtain its prediction result and corresponding certified perturbation size $L_m$.

To study the performance of our method, we use the certified accuracy as the evaluation metric, which is defined as follows.
\begin{equation*}
CA(r) = \frac{\sum_{m=1}^{M}\mathbb{I}(g(\textbf{x}_{\mathcal{G}_m} \oplus \bm{\epsilon})=c)\mathbb{I}(L_m>r)}{M},
\end{equation*}
where $ r \in \mathbb{Z}^+$ represents the certified perturbation size for our graph classifier and $\mathbb{I}$ is an indicator function. The first indicator is to count the number of testing sample that is correctly predicted by our algorithm, and the second indicator is to count the number of testing sample that has its certified perturbation size $L_m$ larger than $r$. Therefore, we use $r$ to represent the certified perturbation size of our graph classifier. To show the performance of our certification method under each specific setting, we calculate the certified accuracy for different $r$ varying from 0 to 16.

\subsection{Experimental Results}
\begin{figure}[htbp]
\centerline{\includegraphics[scale=0.4]{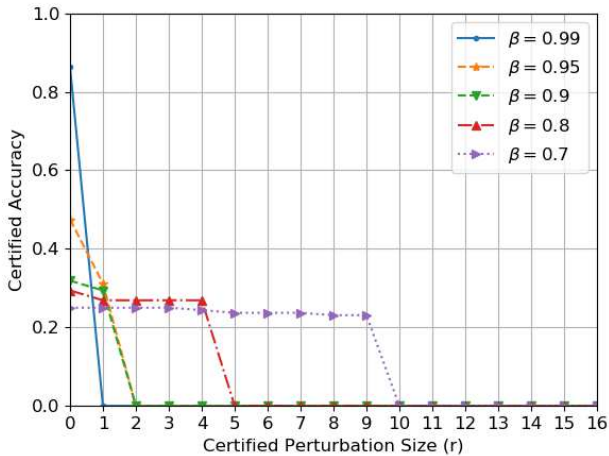}}
\caption{Impact of $\beta$ on our certification method.}
\label{fig:beta}
\end{figure}

We first study the impact of the noise level $(1-\beta)$ on the certified accuracy of our method. Specifically, we show the certified accuracy with respect to the certified perturbation size $r$ under different settings of the noise level. We vary the noise level by changing the value of $\beta$ from 0.7 to 0.99. As shown in Figure~\ref{fig:beta}, we can observe that $\beta$ influences the trade-off between the certified accuracy and radius. We can see that as the value of $\beta$ decreases, the maximum certified perturbation size increases, which means our method is able to certificate the robustness for the graph classifier with higher perturbations, but the certified accuracy decreases. On the other hand, as the value of $\beta$ increases, the maximum certified perturbation size decreases, which means our method is able to certificate the robustness for the graph classifier with higher perturbations, but the certified accuracy increases.


\begin{figure}[htbp]
\centerline{\includegraphics[scale=0.4]{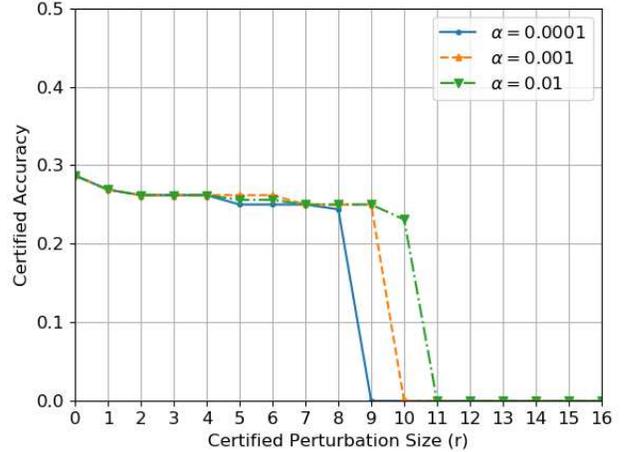}}
\caption{Impact of $\alpha$ on robustness metric.}
\label{fig:alpha}
\end{figure}

Next, we study the impact of confidence level $1-\alpha$ on the performance of our certification method. Specifically, we show the certified accuracy with respect to the certified perturbation size $r$ when the value of $\alpha$ is 0.01, 0.001, and 0.0001, respectively. As shown in Figure~\ref{fig:alpha}, we can see that the confidence level does not have much impact on the certified accuracy when $r$ is small. With the increase of the certified perturbation size $r$, the certified accuracy tends to be influenced by $\alpha$. This is reasonable since smaller $\alpha$ returns a more relaxed estimation of probability. Hence we have a lower bound of certified perturbation size.

Finally, we study the influence of sample time $N$ on the performance of our method by setting $N$ as 1,000, 5,000, and 1,0000, respectively. When the certified perturbation size $r$ is small, the certified accuracy is not influenced by the sample time, since most testing samples have small certified perturbation size, a small sample time $N$ is enough to give a tight estimation of $\underline{p_A}$. With the increase the sample time, the maximum certified perturbation size increases, which means our method tends to certificate the robustness of the graph classifier with higher perturbations. The reason is that larger sample time implies a tighter bound of $\underline{p_A}$, which leads to a high bound of the certified perturbation size.

\begin{figure}[htbp]
\centerline{\includegraphics[scale=0.4]{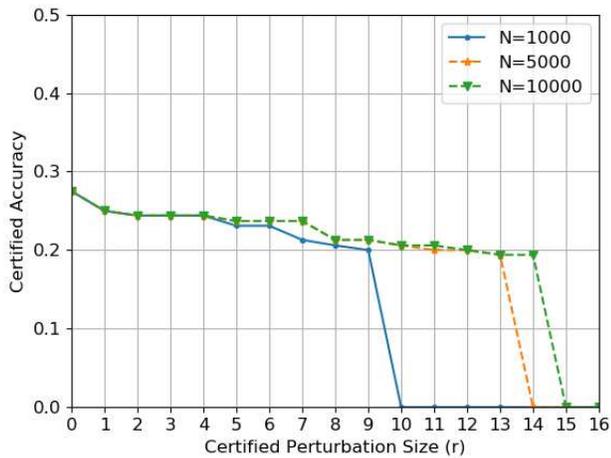}}
\caption{Impact of $N$ on robustness metric.}
\label{fig:N}
\end{figure}

\section{Conclusion}\label{sec:conclusion}
In this work, we have proposed an certification method to build a smoothed graph classification model with certified robustness guarantee. We have proven that the resulting graph classification model would generate the same prediction result for a graph under bounded adversarial perturbations. Our approach could be applied for a wide category of graph classification models. We have conducted extensive experiments on synthetic datasets to demonstrate the effectiveness of our proposed method on a two-layer graph convolutional network (GCN) model. In the future, we will study the certified robustness for other graph-based learning algorithms.


\section*{Acknowledgement}
The work of Z. Gao, R. Hu, and Y. Gong was supported in part by the U.S. National Science Foundation under grants US CNS-2029685 and CNS-1850523. 

\bibliographystyle{IEEEtran}
\bibliography{gong,guo,gao}

\end{document}